\documentclass[letterpaper, 10 pt, conference]{ieeeconf}  

\IEEEoverridecommandlockouts                              

\usepackage{amsmath} 
\usepackage{graphicx}
\usepackage{hyperref}
\usepackage{multirow}
\usepackage{color}
\usepackage{bm}
\usepackage{pifont}
\usepackage{makecell}
\usepackage{booktabs}
\usepackage{stfloats}
\usepackage{algorithm}
\usepackage{algpseudocode}
\usepackage{graphics}

\title{\LARGE \bf
Double-Dot Network for Antipodal Grasp Detection
}

\author{Yao Wang, Yangtao Zheng, Boyang Gao and Di Huang$^{*}$\thanks{Yao Wang, Yangtao Zheng and Di Huang are with the State Key Laboratory of Software Development Environment, School of Computer Science and Engineering, Beihang University, Beijing 100191, China; Boyang Gao is with the Geometry Robotics.}\thanks{*Corresponding author: dhuang@buaa.edu.cn}}

\begin{document}

\maketitle
\thispagestyle{empty}
\pagestyle{empty}

\begin{abstract}
This paper proposes a new deep learning approach to antipodal grasp detection, named Double-Dot Network (DD-Net). It follows the recent anchor-free object detection framework, which does not depend on empirically pre-set anchors and thus allows more generalized and flexible prediction on unseen objects. Specifically, unlike the widely used 5-dimensional rectangle, the gripper configuration is defined as a pair of fingertips. An effective CNN architecture is introduced to localize such fingertips, and with the help of auxiliary centers for refinement, it accurately and robustly infers grasp candidates. Additionally, we design a specialized loss function to measure the quality of grasps, and in contrast to the IoU scores of bounding boxes adopted in object detection, it is more consistent to the grasp detection task. Both the simulation and robotic experiments are executed and state of the art accuracies are achieved, showing that DD-Net is superior to the counterparts in handling unseen objects. 
\end{abstract}
\section{Introduction}

Grasping, an essential and important operation for robotic manipulation, aims to grab and lift objects by grippers using inputs of imaging sensors. For the potential to significantly reduce labor costs, it has been widely attempted and adopted in various applications, such as manufacturing, logistics, and services, in recent years. Unfortunately, grasping is a much more difficult and complicated task to robots than to humans, and it still remains challenge to handle unseen instances of arbitrary poses with diverse textures and shapes.


Robotic grasping has been extensively discussed for years and typically addressed in the 3D space. Early approaches \cite{Nguyen1986Constructing} \cite{Bicchi2000Robotic} \cite{ferrari1992planning} \cite{shimoga1996robot} regard it as a synthesis problem where the input is fitted to the templates (\emph{e.g.} CAD models) whose grasps are assumed available. They only work on known objects, thus impeding its spread in real-world scenarios. To generalize to unknown objects, some approaches employ their 3D point-clouds reconstructed from depth data and estimate grasps by analyzing shape characteristics based on hand-crafted or deep learned geometry features \cite{mousavian20196} \cite{dune2008active} \cite{marton2010general} \cite{kraft2008birth} \cite{popovic2011grasping}. They deliver promising results, but are usually criticized for low running speed. There also exist several studies that consider grasping as a motion planing process and model it by Reinforcement Learning (RL) \cite{levine2018learning} \cite{lampe2013acquiring} \cite{gu2017deep}. While attractive, they generally suffer from long and unstable training phases.

\begin{figure}[t]
	\centering
	\footnotesize
   \begin{tabular}{c@{\hskip3pt}c@{\hskip3pt}c}
	  \includegraphics[width=0.3\columnwidth]{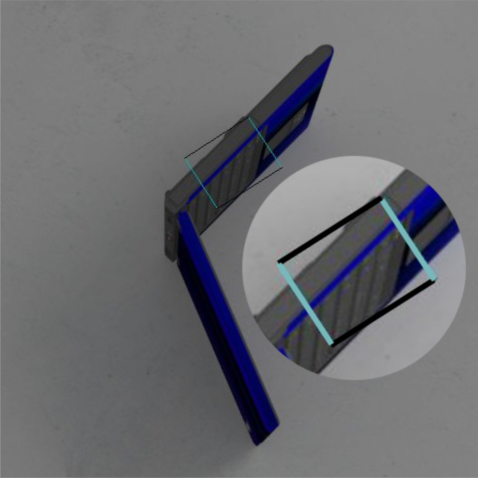}&	
	  \includegraphics[width=0.3\columnwidth]{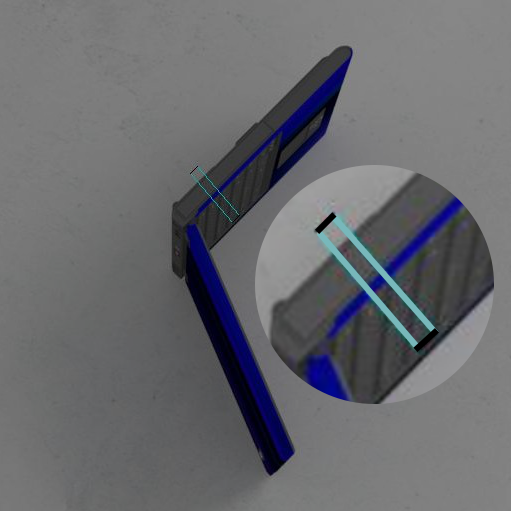}&	
	  \includegraphics[width=0.3\columnwidth]{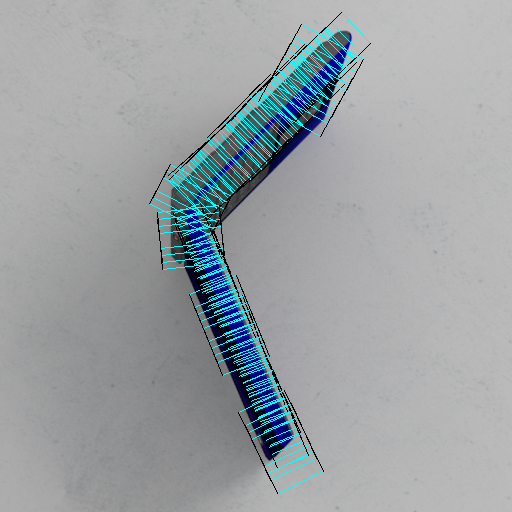}\\
	  \includegraphics[width=0.3\columnwidth]{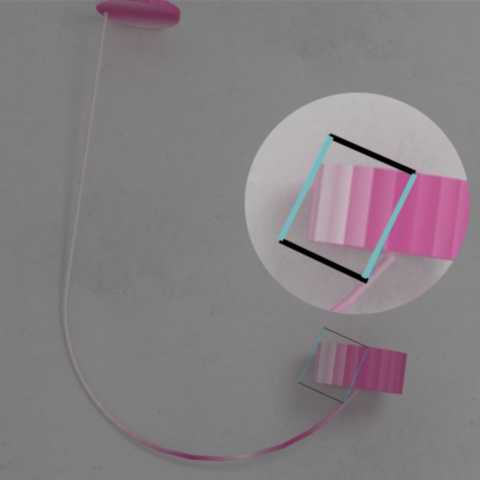}&
	  \includegraphics[width=0.3\columnwidth]{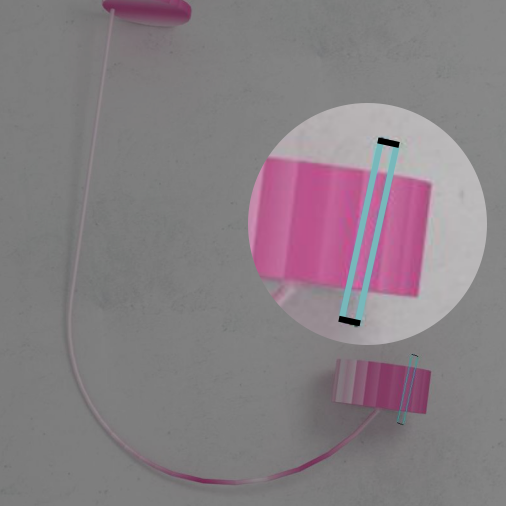}&
	  \includegraphics[width=0.3\columnwidth]{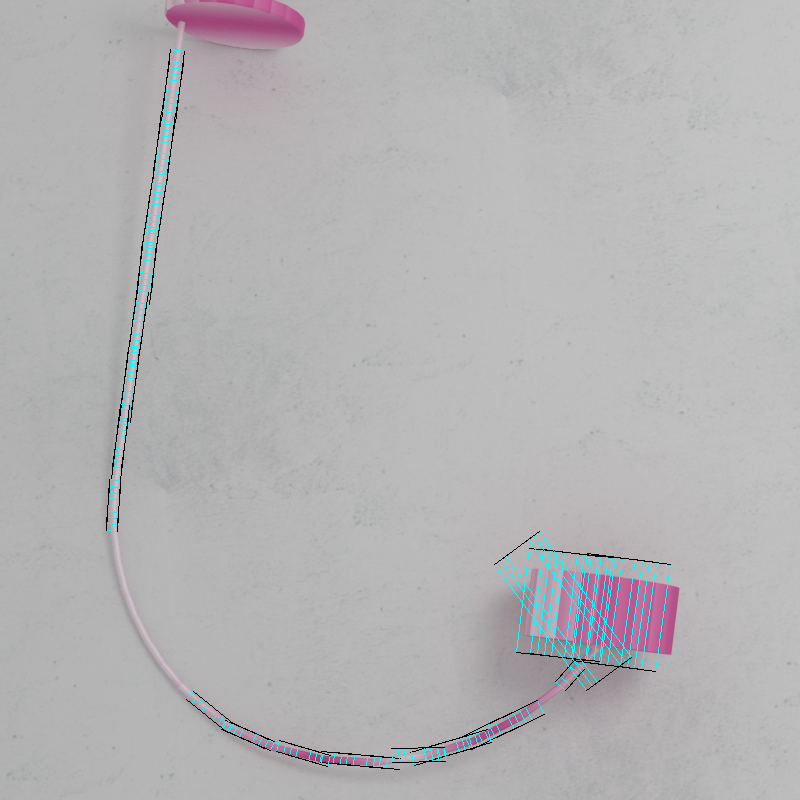}\\
	  \includegraphics[width=0.3\columnwidth]{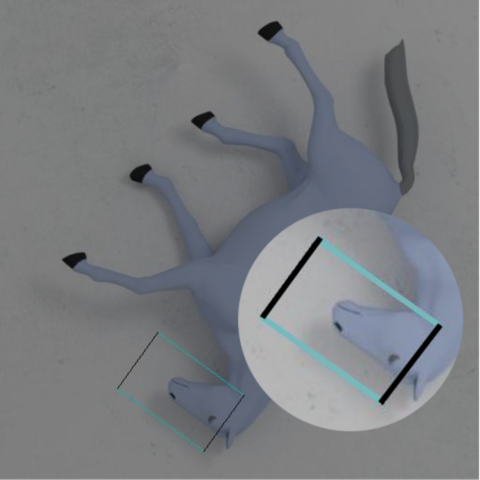}&
	  \includegraphics[width=0.3\columnwidth]{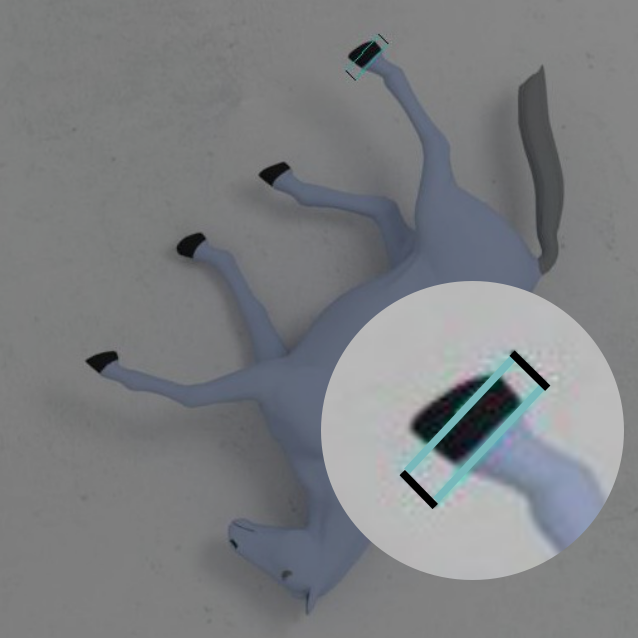}&	
	  \includegraphics[width=0.3\columnwidth]{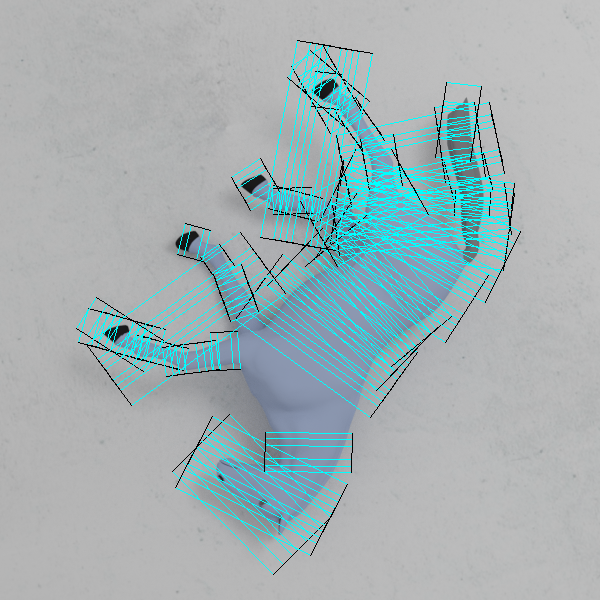}\\
		(a) \cite{zhou2018fully} & (b) Ours & (c) Ground Truth\\
	\end{tabular}
	\caption{Visualization of predictions and ground-truths. (a) and (b) show the grasp results achieved by \cite{zhou2018fully} in bounding boxes and the proposed method in double-dots, respectively, and (c) shows the corresponding ground-truths.}\label{fig:figure0}
\end{figure}
Recently, due to the great demand in applications, robotic grasping is simplified to the 2D domain, \emph{i.e.} views and lifts are perpendicular to support surfaces. In this way, the searching space of grasps becomes smaller and the acquirement of successful grasps becomes easier. Compared to the investigations that adapt the RL idea from 3D to 2D \cite{breyer2018flexible} \cite{mahler2017learning}, much more efforts are made to the detection framework, where the successful grasp is treated as a special kind of targets within the given image, and it samples grasps and determines their validity. The sampling efficiency and classification accuracy of the preliminary detection methods are limited by the hand-designed features \cite{chinellato2003ranking} \cite{mahler2017dex} \cite{bowers2003manipulation} \cite{morales2004learning}. Such drawbacks are overcome in the descendant alternatives through deep CNN features and end-to-end learning mechanisms. For example, \cite{jiang2011efficient} \cite{lenz2015deep} represent a planar grasp as a 5-dimensional gripper configuration, formed by an oriented rectangle, and apply the advanced object detection pipeline that first generate the anchors, then simultaneously regress and evaluate the grasp candidates, and finally select the best grasp configuration. 

In fact, object detection and grasp detection are not exactly the same, and direct extension from the former to the latter does not well address their differences, thereby leaving room for improvement. As in \cite{mahler2017dex} \cite{wang2016robot} \cite{xu2019graspcnn}, they modify the gripper configuration in terms of dimensionality or pattern so that it better accommodates to grasp detection in different data modalities (\emph{i.e.} RGB images or depth maps). However, the gap is not eliminated. As we know, object detection methods \cite{ren2015faster} \cite{liu2016ssd} \cite{liu2018receptive} commonly exploit Intersection over Union (IoU) to evaluate the quality of the returned rectangle (bigger IoU is better), but this measurement is too coarse for grasp detection. Even though a grasp rectangle has a good IoU, a small error in rotation or translation may force the gripper to collide with the object, incurring execution failure, as in Fig.~\ref{fig:figure0}. More importantly, current state of the art detectors are anchor-based and their accuracies are sensitive to pre-defined anchor scales and aspect ratios, which makes them problematic to generalize to unseen objects with large shape differences.

In this paper, we propose a new deep learning approach, namely Double-Dot Network (DD-Net), to antipodal robotic grasping, which dedicates to alleviate the gap between object detection and grasp detection. It follows the anchor-free concept of CornerNet \cite{law2018cornernet} and CenterNet \cite{duan2019centernet} that detects an object as very few keypoints, which does not require empirically pre-set anchors, thus allowing more generalized and flexible prediction on unseen objects. Specifically, in contrast to current detection based grasping methods delivering bounding boxes (or other shapes, \emph{e.g.} circles \cite{xu2019graspcnn}), DD-Net generates the gripper configuration as a pair of fingertips. An effective CNN architecture is introduced to localize such fingertips in a one-stage manner, and with the help of auxiliary centers for refinement, DD-Net accurately and robustly infers grasp candidates. In addition, a loss function is designed to evaluate grasps, which measures the quality of the fingertips rather than the IoU scores of the bounding boxes, making the optimization consistent to the task. Simulation experiments are executed on the major public grasping platform, Jacquard, reporting the state of the art SGT accuracy and very competitive results are also achieved on robotic experiments compared to the counterparts in terms of accuracy and efficiency.

\section{Related Work}
In the literature, planar grasping is generally formulated in two ways. One is motion planning, mainly handled by RL. Breyer \emph{et al.}~\cite{breyer2018flexible} model this problem as a Markov Decision Process, where a neural network is designed to predict the actions and finger joint positions and Trust Region Policy Optimization is used to update policy parameters. 



Another is detection, where candidate sampling and graspability scoring are conducted in series. To generate grasp proposals, Le \emph{et al.}~\cite{le2010learning} represent contact points by a set of hand-crafted features, including depth and pixel intensity variations, gradient angles, distances, \emph{etc.}. They further evaluate those candidates by an SVM classifier. Mahler~\emph{et al.} \cite{mahler2017dex} select the grasp by iteratively sampling a set of candidates and re-fitting the grasp distribution according to the grasp quality given by a pre-trained deep neural network (\emph{i.e.} GQ-CNN) based scorer. This method shows promising results, but suffers from a low running speed. Gariépy \emph{et al.} \cite{gariepy2019gq} train an STN (Spatial Transformer Network) to accelerate sampling; however, STN is supervised by GQ-CNN, which is not accurate enough, thus limiting the performance.


\begin{figure}[h]    
    \vspace{4pt}                              
	\centering                                     
	\includegraphics[width=0.7\columnwidth]{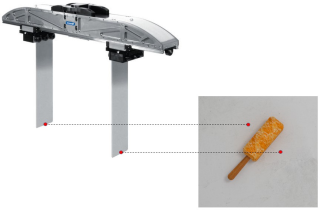}   
	\caption{Illustration of double-dot grasps. Dots marked in red are fingertips of a parallel-jaw gripper.}                                 
	\label{fig:grasp}                               
\end{figure}

 
More recent studies follow the object detection pipeline. \cite{guo2017hybrid} \cite{zhou2018fully} \cite{chu2018real} \cite{karaoguz2019object} adapt the two-stage detectors, such as Faster-RCNN \cite{ren2015faster} and RRPN \cite{ma2018arbitrary} to grasp detection. Considering the difference between the two tasks, the major issue is to predict the orientation of the bounding box. \cite{guo2017hybrid} \cite{chu2018real} discretize the angle and convert its prediction to a classification problem, while \cite{zhou2018fully} \cite{karaoguz2019object} employ anchor boxes with different orientations. Those solutions indeed take a step to grasp detection, but due to the inheritance of IoU based optimization and anchor based mechanism in object detection, there exists much space for improvement. \cite{zhou2018fully} \cite{chen2019convolutional} replace IoU minimization by point metric or grasp path; unfortunately, the methods are still too rough. \cite{asif2018graspnet} \cite{ghazaei2018dealing} attempt to generate better grasp configuration from pixel-wise prediction. While attractive, their models are very preliminary.

\section{Problem Formulation}
\label{sec: pf}

In this work, we target the task of planar grasping, which localizes two fingertips of the parallel-jaw gripper equipped on the robotic arm in the image coordinates where it has the maximum possibility to successfully lift the object laying on the plane. 

Formally, a grasp is defined as double-dot representation:

\begin{equation}
\bm{g}=\left\{ \bm{c}_1=(x_1, y_1), \bm{c}_2=(x_2, y_2)\right\} \,,
\end{equation}

\noindent where $ \bm{c}_i$ denotes the center of a fingertip of a parallel-jaw gripper. Fig.~\ref{fig:grasp} illustrates the fingertips relative to the object. 

\begin{figure*}[t]              
    \vspace{4pt}                    
	\centering                                     
	\includegraphics[width=0.9\textwidth]{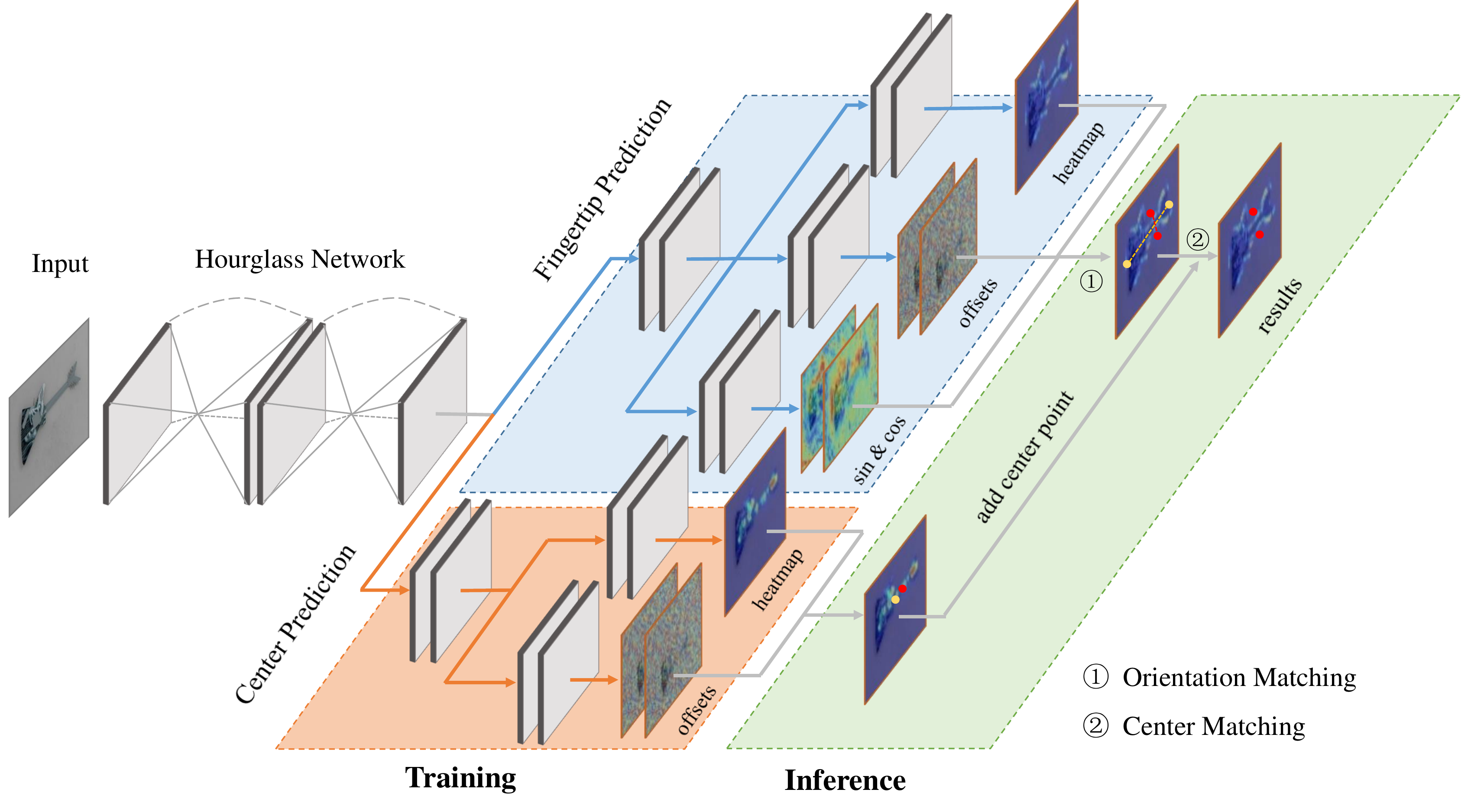}   
	\caption{Architecture of DD-Net. It consists of three major parts: an Hourglass network, a fingertip prediction branch and a center prediction branch. Orientation matching and center matching only work for inference. }                                 
	\label{fig:train}                           
\end{figure*}
\section{Method}


\subsection{Grasp Detection Revisit}

Recently, a number of object detection methods have been adapted to grasp detection, but as displayed in Fig. \ref{fig:figure0}, correct detection does not always guarantee a success of grasp. This gap between object detection and grasp detection is mainly caused by the definition of grasp representation.

As we know, in the majority of the existing public datasets, to match the concept of the anchor box in object detection, a grasps is annotated as a 5-dimensional rectangle:

\begin{equation}
\bm{g}^{\prime}=\left\{x, y, h, w, \theta\right\}\,,
\end{equation}

\noindent where $(x, y)$ is the center of the rectangle; $w$ and $h$ represent the width and the height of the rectangle, \emph{i.e.}, the opening size and the jaw size of the end effector; and $\theta$ denotes the orientation of the rectangle with regard to the horizontal axis. As the jaw size is fixed, $h$ can be ignored in practice.

The grasp quality is decided by the Intersection over Union (IoU) between the detected rectangle and the corresponding Ground-Truth (GT), where a bigger IoU indicates a better grasp. However, such a measurement is too coarse in grasp detection, as a small error of the grasp rectangle in terms of rotation or translation tends to make the gripper touch the object, leading to execution failure, although it still holds a good IoU. On the other side, anchor-based detectors currently dominate the community, but their accuracies are sensitive to the anchor setting, \emph{i.e.} scales as well as aspect ratios. This mechanism is problematic in grasp detection as it is expected to handle unseen objects even with big differences in shape.

\begin{figure}[h]
	\centering
	\footnotesize
   \begin{tabular}{c@{\hskip2pt}c}
      \includegraphics[width=0.45\columnwidth]{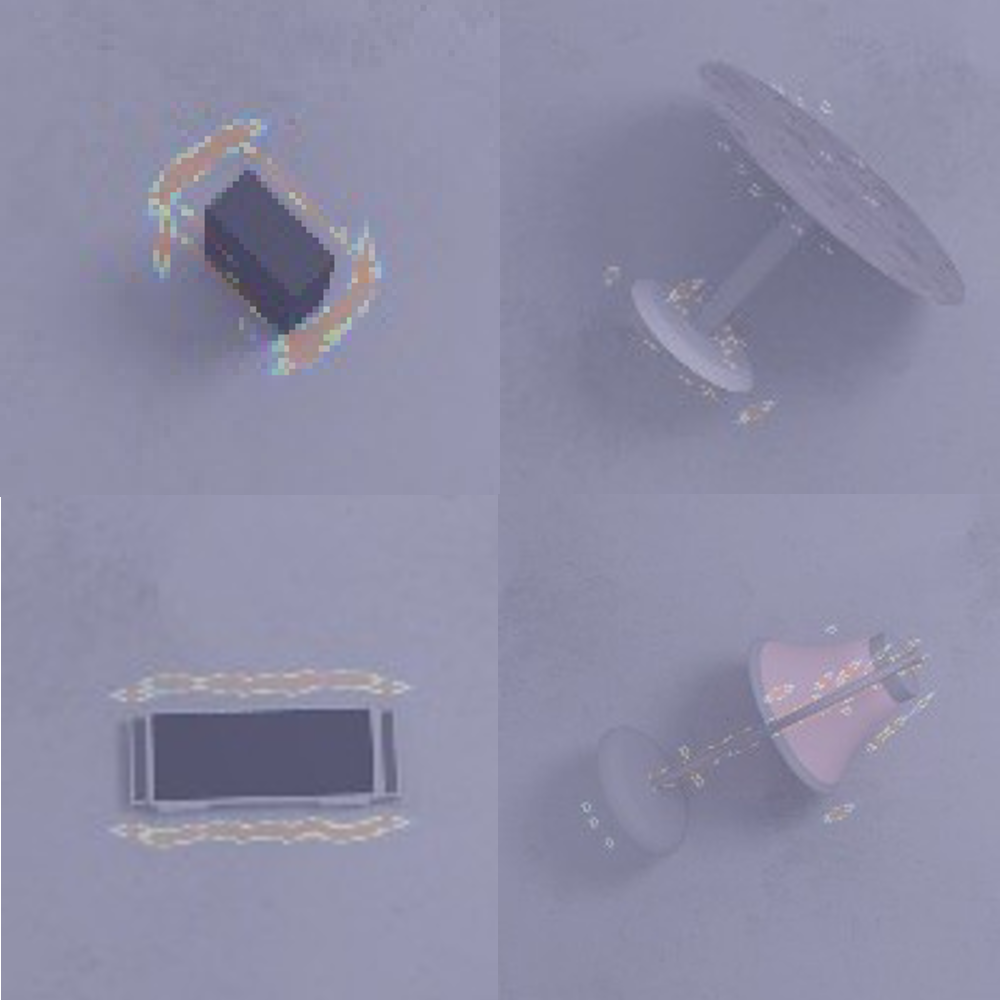}&%
      \includegraphics[width=0.45\columnwidth]{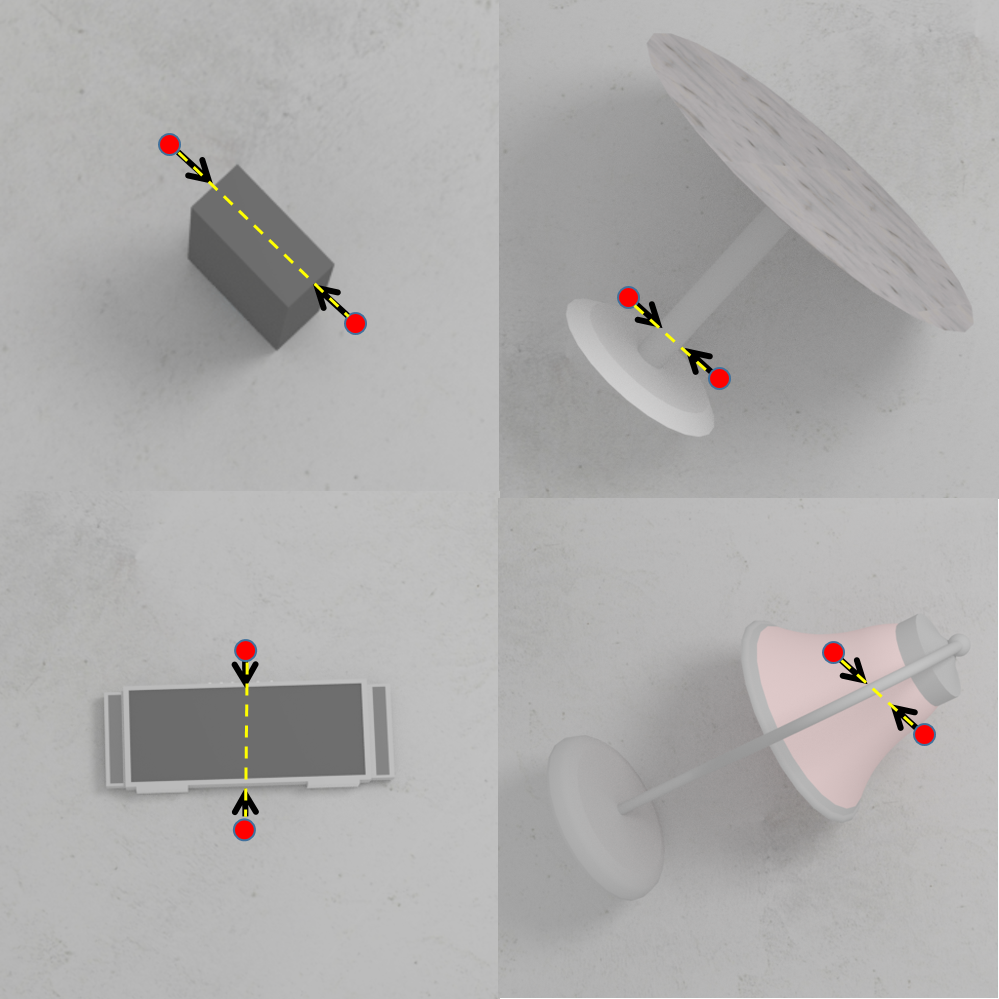}\\
		(a) Double-dot labels   & (b) Orientation labels \\
	\end{tabular}
	\caption{Samples: (a) to increase double-dot labels, pixels around GT fingertips are annotated according to an unnormalized 2D Gaussian distribution and (b) the orientation is defined as the angle towards the center point.}
	\label{fig:label}
\end{figure}

The analysis above motivates us to re-define the grasp by double-dot representation as in Sec. \ref{sec: pf} and develop anchor-free grasp detectors.

\subsection{Data Labelling}

To make full use of existing datasets, where the labels are provided as the common rectangle representation, for model building, we convert rectangle representations to double-dots representation as in Sec. \ref{sec: pf}.




Considering that rectangle labels are sparse, we assume that the locations nearer to the GT fingertips converted from the bounding boxes are more likely to be positive samples, which increases double-dot labels. In this case, we label the pixel $\bm{p}$ around GT fingertips according to an unnormalized 2D Gaussian distribution, which is defined as:

\begin{equation}
	s(\bm{p}) = \min \left ( 1, \sum_{g \in \boldsymbol{G}} \sum_{i=1,2} \frac{\exp(-Q R(\theta) \Sigma^{-1} R^{T}(\theta) Q)}{\sigma^{(g)}_{x} \sigma^{(g)}_{y}} \right ) \,,
\end{equation}

\noindent where $Q = \bm{p}-\bm{c}^{(g)}_{i}$, $\Sigma = diag(\sigma^{(g)}_{x}, \sigma^{(g)}_{y})^{2}$, $R(\theta)$ is the rotation matrix and $\boldsymbol{G}$ is the ensemble of grasp annotations within an image. In this study, $\sigma^{(g)}_{x}$ and $\sigma^{(g)}_{y}$ are empirically set as 1 and 0.75$h$ respectively so that fingertips are not on objects. 

The orientation of each fingertip $\theta_{\bm{c}_i}$ is defined as the angle towards the center point:

\begin{equation}
	\theta_{\bm{c}_i}=\arctan\frac{(y_i-y_{i}^{center})}{(x_i-x_{i}^{center})} \,.
\end{equation}

The illustration of the fingertips and corresponding orientations are shown in Fig.~\ref{fig:label}.

\subsection{Network Architecture} 

Inspired by the latest advances in general object detection, an object can be detected as very few (two or three) fiducial points \cite{law2018cornernet} \cite{duan2019centernet}, we propose a novel single-stage approach to grasp detection, namely Double-Dot Network (DD-Net). Instead of predicting the traditional 5-dim grasp rectangle, we directly detect two fingertips as a grasp. Meanwhile, as the grasp is more stable when it is nearer to the center of the object, we also localize the gripper center as an auxiliary point for refinement. 


Specifically, DD-Net includes three key components: image encoding, fingertip prediction and center prediction, as shown in Fig.~\ref{fig:train}. We adopt an off-the-shelf convolutional neural network, \emph{i.e.} Hourglass~\cite{newell2016stacked}, to extract image features, conveying local and global clues for comprehensive description. These features then enter into two branches: one for fingertip prediction and another for center prediction. In the fingertip branch, we compute the score, offset and orientation, while in the center branch, we only calculate the score and offset. The focal loss $L_{det}$ is used to train the network to score for each pixel to be a fingertip or a center point, defined as:

\begin{small}
\begin{equation}
L_{det} = -\sum \limits_{h, w} 
\begin{cases}
(1 - q_{\bm{p}})^{\alpha} \log (q_{\bm{p}}), & \text{if $s_{\bm{p}}=1$} \\[2ex]
(1 - s_{\bm{p}})^{\beta}(q_{\bm{p}})^{\alpha} \log(1 - q_{\bm{p}}), & \text{otherwise}
\end{cases} \,,
\end{equation}
\end{small}

\noindent where $q_{\bm{p}}$ and $s_{\bm{p}}$ are the predicted score and the GT label at pixel $\bm{p}$, respectively. $\alpha$ and $\beta$ are the hyper-parameters that weight the contribution of each pixel.

To remap the locations from heatmaps to the input image, an offset is predicted. The offset is defined as 
$\boldsymbol{o}_g = (\frac{x_{g}}{n}-\lfloor\frac{x_{g}}{n}\rfloor, \frac{y_{g}}{n}-\lfloor\frac{y_{g}}{n}\rfloor)$
where $n$ is the down-sampling factor of the network. The offset loss is defined as

\begin{equation}
L_{off} = \sum_{g \in \boldsymbol{G}} SmoothL1 \left( \boldsymbol{o}_g, \boldsymbol{\hat{o}}_g \right) \,,
\end{equation}

To predict the orientation of the plate, we apply the smooth L1 Loss to regress the $\sin$ and $\cos$ value of the angle.

\begin{equation}
	L_{ori}^{sin} =  \sum_{\boldsymbol{g} \in \boldsymbol{G}} \sum_{i=1,2} SmoothL1 \left( \eta, \sin(\theta_{\bm{c}_{i}^{(g)}}) \right) \,,
\end{equation}

\noindent where $\eta$ is the $\sin$ value predicted by the network. $L_{ori}^{cos}$ can be similarly defined.

We define our final loss function as follows:
\begin{equation}
\begin{aligned}
L = L_{det}^{con}+L_{det}^{cen}+L_{off}^{con}+L_{off}^{cen}+L_{ori}^{cos}+L_{ori}^{sin}
\end{aligned}
\end{equation}

\noindent where $L_{det}^{con}$ and $L_{det}^{cen}$ denote the focal losses, which are used to train the network to detect fingertips and center points respectively. $L_{off}^{con}$ and $L_{off}^{cen}$ are used to train the network to predict their offsets. $L_{ori}^{cos}$ and $L_{ori}^{sin}$ are used to train the network to predict the plate orientation of each fingertip.

\subsection{Fingertip Grouping}

\begin{figure}[h]
	\centering
	\footnotesize
   \begin{tabular}{c@{\hskip4pt}c}
      \includegraphics[width=0.47\columnwidth]{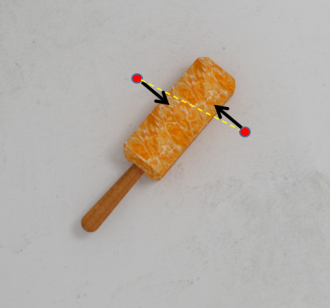}&%
      \includegraphics[width=0.47\columnwidth]{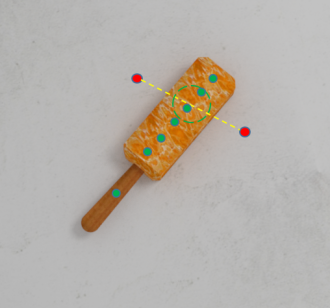}\\
		(a) Orientation Matching  & (b) Center Matching \\
	\end{tabular}
	\caption{Two principles to filter candidates: (a) orientation matching accepts the paired fingertips if the angle is below a threshold and (b) center matching accepts the ones if the central region contains centers (green points).}                          
	\label{fig:grouping}  
\end{figure}

During inference (see Algorithm. \ref{alg:inf}), we first sample the fingertip candidates from the heatmaps generated by the network and then group them for refinement. Two grouping principles, \emph{i.e.}, orientation matching and center matching, are designed to filter the candidates, illustrated in Fig.~\ref{fig:grouping}. 

Orientation matching groups fingertips with the prediction of the plate orientation and the non-antipodal pairs are rejected. A pair of fingertips is filtered, if one of the predicted orientation of a fingertip deviates that of the fingertip pair over a threshold. Center matching defines a central region as a circle centered at the midpoint of the two fingertips, similar to CenterNet \cite{duan2019centernet}. If the central region of a fingertip pair does not contain any center point, the fingertip pair is rejected. In our experiments, the radius of a central region is set as $\frac{1}{3}$ of the distance between fingertips.

After filtering the candidates, the score of the remaining grasp is defined as the sum of the ones of the three points, \emph{i.e.} two fingertips and its center. Finally, we select the grasp with the highest score as the final prediction to execute. 

\begin{algorithm}[h]
	\caption{Inference process} 
	\hspace*{0.02in} {\bf Input:} 
	input image $I$\\
	\hspace*{0.02in} {\bf Output:} 
	grasp prediction ${\bm{g}^{*}}$
	\begin{algorithmic}[1]
	\State Generate heatmap, offset and orientation of fingertips by DD-Net. 
	\State Generate heatmap and offset of centers by DD-Net.
	\State Select top-$k$ fingertip and center candidates on heatmaps.
	\State Adjust their positions with the predicted offsets. 
	\State Group $k$ fingertips as grasp candidates and filter them using the orientation and center matching strategies.
	\State Compute the grasp score by the sum of the three points. Rank and select the grasp with the highest score as final prediction ${\bm{g}^{*}}$.
	\State \Return ${\bm{g}^{*}}$
	\end{algorithmic}
	\label{alg:inf}
\end{algorithm}

\section{Experiments}

To evaluate the proposed method, we carry out both simulation and robotic experiments. The databases, metrics, protocols, implementation details, and results are described in the subsequent.

\subsection{Datasets and Metrics}

The simulation experiments are conducted on the Jacquard and Cornell benchmarks and the robotic ones are on a  self-built database.


The Jacquard dataset contains 54k images of 11k objects with more than one million unique grasp locations labeled by simulation trials. It supports the evaluation by Simulated Grasp Trails (SGT). Depierre \emph{et al.}~\cite{depierre2018jacquard} release a web interface where the scene index and the corresponding grasp prediction can be uploaded. The scene is rebuilt in the simulation environment and the grasp is performed by the simulated robot. A grasp is considered successful if the object is correctly lifted, moved away and dropped at a given location.

The Cornell Grasp dataset is one of the most common datasets used in grasp detection. It includes 885 images, and each image is labeled by a few graspable rectangles. The evaluation is based on the rectangle metric with the standard threshold values of $30^{\circ}$ for the angle and 25\% for IoU. 

\begin{figure}[h]
	\centering
	\footnotesize
   \begin{tabular}{c@{\hskip2pt}c}
      \includegraphics[width=0.4\columnwidth]{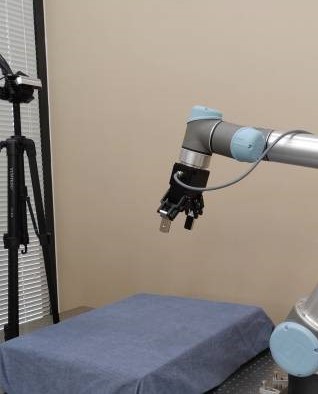}&%
	  \includegraphics[width=0.5\columnwidth]{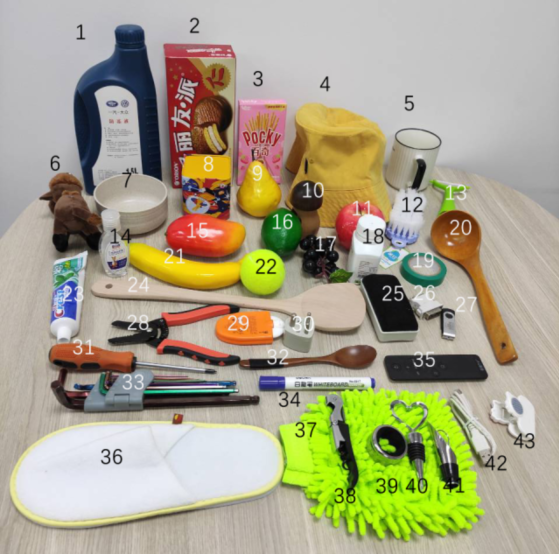}\\
		(a) hardware set-up  & (b) objects \\
	\end{tabular}
	\caption{Device and dataset of robotic experiments}
	\label{fig:setup_objects}
\end{figure}

The self-built dataset consists of 43 objects with different textures, shapes, materials, and scales, as shown in  Fig. ~\ref{fig:setup_objects}. Success rate is used as the evaluation metric.

\subsection{Protocols}

Both the Jacquard and Cornell datasets are divided into a training set and a testing set with a ratio of 4:1. There are two ways to split the dataset: image-wise split and object-wise split. Image-wise split means the training and testing sets are divided randomly, while object-wise split means objects in the test set do not appear in training. For fair comparison, as in previous work, we split the Jacquard dataset by object-wise and the Cornell Grasp dataset both by image-wise and by object-wise. 

To study the generalization ability on unknown objects in the real-world case, we directly deploy DD-Net (trained on Jacquard without fine-tuning) to an UR5 collaborative robot with a Backyard E140F gripper and a fixed Intel RealSense D435 camera. We coarsely estimate the height by averaging the values around the fingertip position predicted in the depth channel. All the experiments are launched on a desktop with an AMD Ryzen 5 2600 Six-Core Processor and a graphics card of NVIDIA GeForce 1080. Each object is placed in three different stable poses on a table. A grasp is considered successful if the robot lifts the object and drops it to the appointed area. 

\begin{table}[!ht]
	\centering
    \vspace{7pt}
	\caption{Comparison with state of the arts in terms of SGT and rectangle metric by object-wise on the Jacquard database.}
	\label{tab:tab2}
    \centering 
	\begin{tabular}{l|c|c}
    \hline
	 \multirow{2}{*}{\textbf{Algorithm}} & \multicolumn{2}{c}{\textbf{Accuracy (\%)}} \\
	 \cline{2-3}
	 & \textbf{SGT} & \textbf{Rectangle} \\
	  \hline
	 Depierre \emph{et al.} \cite{depierre2018jacquard} & 72.42 &74.2 \\ 
	 Zhou  \emph{et al.}\cite{zhou2018fully} & 81.95  & 92.8\\
	 Zhang  \emph{et al.}\cite{zhang2019roi} & -  & 93.6\\
	 Depierre \emph{et al.}\cite{depierre2020optimizing} &85.74& - \\
     Kumra \emph{et al.} \cite{kumra2020antipodal} &-& 94.6 \\
    \hline
	 \textbf{DD-Net}  & \textbf{89.41} &\textbf{97.0}\\
	 \hline
	\end{tabular}
\end{table}

\begin{table}[h]
	\centering
	\caption{Comparison in terms of rectangle metric by image-wise (IW) and object-wise (OW) on the Cornell Grasp database.}
	\label{tab:tab1}
	\begin{tabular}{l|c|c|c}
	\hline
	\multirow{2}{*}{\textbf{Algorithm}} & \multirow{2}{*}{\textbf{Input}} & \multicolumn{2}{c}{\textbf{Accuracy (\%)}} \\
	\cline{3-4}
	& & \textbf{IW} & \textbf{OW} \\
	\hline
	SAE, struct. reg. Two stage \cite{lenz2015deep}& RGB-D & 73.9 & 75.6 \\
	MultiGrasp \cite{redmon2015real}& RGB-D & 88.0 & 87.1 \\
	Multi-model \cite{kumra2017robotic}& RGB & 88.8 & 87.7 \\
	ZF-net \cite{guo2017hybrid}& RGB-D & 93.2 & 89.1 \\
	STEM-CaRFs \cite{asif2017rgb}& RGB-D & 88.2 & 87.5 \\
	GraspPath \cite{chen2019convolutional}& RGD & 86.4 & 84.7\\
	GraspNet \cite{asif2018graspnet}& RGB-D & 90.6 & 90.2 \\
	Multiple, M=10 \cite{ghazaei2018dealing}& RGB & 91.5 & 90.1 \\
	VGG-16 model \cite{chu2018real}& RGB-D & 95.5 & 91.7 \\
	ResNet-50 model \cite{chu2018real}& RGB-D & 96.0 & 96.1 \\
	ResNet-50 FCGN \cite{zhou2018fully}& RGB & \textbf{97.7} & 94.9 \\
	ResNet-101 FCGN \cite{zhou2018fully}& RGB & \textbf{97.7} & \textbf{96.6} \\
	ROI-GD, ResNet-101 \cite{zhang2019roi}& RGB & 93.6 & 93.5 \\
	ROI-GD, ResNet-101 \cite{zhang2019roi}& RGD & 92.3 & 91.7 \\
    GR-ConvNet \cite{kumra2020antipodal}&RGB& 96.6&95.5\\
    GR-ConvNet \cite{kumra2020antipodal}&RGBD& 97.7&96.6\\
	\hline
    \textbf{DD-Net, ResNet-50} & RGB & 96.1 & 95.5\\
    \textbf{DD-Net, ResNet-101} & RGB & 96.6 & 96.0\\
	\textbf{DD-Net, Hourglass-104} & RGB & 97.2 & 96.1\\
	\hline
	\end{tabular}
\end{table}

\begin{figure}[!h]
   \begin{tabular}{c@{\hskip3pt}c}
      \includegraphics[width=0.47\columnwidth]{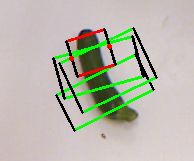}&%
	  \includegraphics[width=0.47\columnwidth]{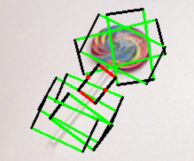}\\
	\end{tabular}
	\caption{Good grasps (red rectangles) by DD-Net are misjudged by Cornell GTs (green rectangles).}
	\label{fig:cornell_vis}
\end{figure}

\begin{table*}[!tp]
	\centering
    \vspace{7pt}
	\caption{Results of robotic experiment. The object index is shown in Fig. ~\ref{fig:setup_objects}.}
	\label{tab:tab13}
	\begin{tabular}{c|c|c|c|c|c|c|c|c|c|c|c|c|c|c|c}
		\toprule
		Object index & \#1 & \#2 & \#3 & \#4 & \#5 & \#6 & \#7 & \#8 & \#9 & \#10  & \#11 & \#12 & \#13 & \#14 & \#15 \\
		DD-Net & 2/3 & 3/3 & 3/3 & 3/3 & 3/3 & 2/3 & 3/3 & 3/3 & 3/3 & 3/3 & 3/3 & 2/3 & 3/3 & 3/3 & 3/3 \\
		Dex-Net & 2/3 & 2/3 & 3/3 & 3/3 & 2/3 & 3/3 & 0/3 & 2/3 & 2/3 & 3/3 & 2/3 & 3/3 & 2/3 & 3/3 & 2/3 \\
		\midrule
		\midrule
		Object index & \#16 & \#17 & \#18 & \#19 & \#20 & \#21 & \#22 & \#23 & \#24 & \#25 & \#26 & \#27 & \#28 & \#29 & \#30  \\
		DD-Net & 3/3 & 3/3 & 2/3 & 2/3 & 3/3 & 3/3 & 2/3 & 3/3 & 3/3 & 3/3 & 3/3 & 3/3 & 3/3 & 3/3 & 3/3 \\
		Dex-Net & 2/3 & 3/3 & 3/3 & 2/3 & 3/3 & 3/3 & 2/3 & 3/3 & 2/3 & 3/3 & 3/3 & 2/3 & 1/3 & 1/3 & 3/3 \\
		\midrule
		\midrule
		Object index & \#31 & \#32 & \#33 & \#34 & \#35 & \#36 & \#37 & \#38 & \#39 & \#40 & \#41 & \#42 & \#43 & \multicolumn{2}{|c}{Overall}  \\
		DD-Net & 3/3 & 2/3 & 2/3 & 3/3 & 3/3 & 3/3 & 2/3 & 3/3 & 3/3 & 3/3 & 3/3 & 3/3 & 3/3 & \multicolumn{2}{|c}{93.0\%} \\
		Dex-Net & 2/3 & 0/3 & 1/3 & 2/3 & 2/3 & 3/3 & 3/3 & 2/3 & 2/3 & 2/3 & 2/3 & 3/3 & 3/3 & \multicolumn{2}{|c}{75.2\%} \\
		\bottomrule
	\end{tabular}
\end{table*}


\subsection{Implementation Details}
DD-Net is implemented in PyTorch, with the training and inference accomplished on GTX1080Ti GPUs. 

In training, the model is randomly initialized under the default setting without pre-training on any external dataset. The resolution of the input image is 511$\times $511 and that of the output is 128$\times $128. The images in Jacquard are directly resized to 511$\times $511. Due to the difference in aspect ratio, the images in Cornell are first cropped to 480$\times $480 and then rescaled to 511$\times $511. The Adam algorithm is used for optimization. The learning rate is set at 2.5$\times $10$^{-4}$, and the training mini-batch size is set as 10. During training, we conduct data augmentation including random horizontal flipping, random color jittering, and random rotating to avoid over-fitting. 

In inference, we take the top 70 fingertips suggested by the heatmaps. The grasp with an opening size greater than 70 pixels or smaller than 2 pixels is rejected so that it is executable. To make the results compatible to those using 5-dim parameters to represent grasp configurations, the plate length is set at 10 pixels and 35 pixels for Jacquard and Cornell respectively.

\subsection{Evaluation on Jacquard}

\autoref{tab:tab2} summarizes the results of different methods on Jacquard in terms of both the criteria of SGT and rectangle metric. It can be seen that the proposed DD-Net outperforms the state of the art counterparts \cite{depierre2018jacquard} \cite{zhou2018fully} \cite{depierre2020optimizing} by a large margin, with an improvement of 3.7\% in SGT using RGD images as input, which demonstrates its superiority in simulation evaluation. Meanwhile, we can also notice that DD-Net reaches the best score using the standard rectangle metric, with a gain of 2.4\% over the second best reported by \cite{kumra2020antipodal}, showing its good compatibility. Finally, as the experiments are conducted in an object-wise way, this performance clearly indicates the effectiveness of DD-Net in handling unseen objects.

\subsection{Evaluation on Cornell}
\autoref{tab:tab1} displays the comparative results using the rectangle metric on the Cornell Grasp database. We can see that DD-Net delivers very competitive scores, comparable to the top ones in both the image-wise and object-wise scenarios. The reason that it does not show obvious increase in accuracy lies in that the GT annotations on Cornell are not as complete as those in Jacquard and some good grasps are not correctly judged as shown in Fig.~\ref{fig:cornell_vis}. On the other side, we evaluate the proposed approach with different backbones, and it can be seen that DD-Net with the light and popular ResNet-50 still reaches good results and that with stronger backbones, \emph{i.e.} ResNet-101 and Hourglass-104, slightly increase the performance.

\subsection{Robotic Evaluation}
The results of robotic experiments are displayed in \autoref{tab:tab13}, where DD-Net demonstrates its strong ability in dealing with unseen objects. Refer to the video attached for more examples. We take the reputed open-source Dex-Net as the counterpart and we can see that DD-Net significantly outperforms Dex-Net by 17.8\% in success rate. Meanwhile, its planning time is 4 times faster (0.26s \emph{vs.} 1.0s).
	
	

\begin{table}[!h]
	\centering
	\caption{Ablation study of orientation matching and center matching in fingertip grouping using SGT on Jacquard.}
	\label{tab:tab12}
	\begin{tabular}{c|c|c}
	 \hline
	 \makecell[c]{\textbf{Orien. Match.}} & \makecell[c]{\textbf{Cent. Match.}} & \textbf{Accuracy (\%)} \\
	 \hline
  \ding{55} & \ding{55} & 52.5 \\
  \hline
  \ding{55} & \ding{51} & 55.1 \\
  \hline
  \ding{51} & \ding{55} & 85.6 \\
	 \hline
  \ding{51} & \ding{51} & 89.4 \\
	 \hline
	\end{tabular}
\end{table}
\begin{figure}[h]
	\centering
	\footnotesize
   \begin{tabular}{c@{\hskip3pt}c@{\hskip3pt}c}
	  \includegraphics[width=0.3\columnwidth]{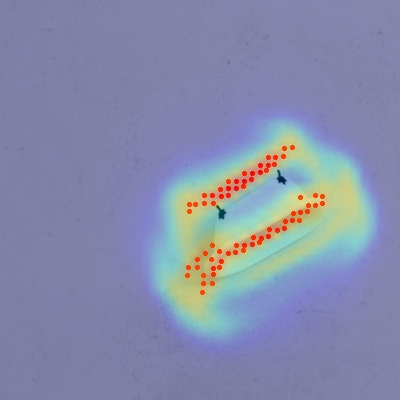}&%
	  \includegraphics[width=0.3\columnwidth]{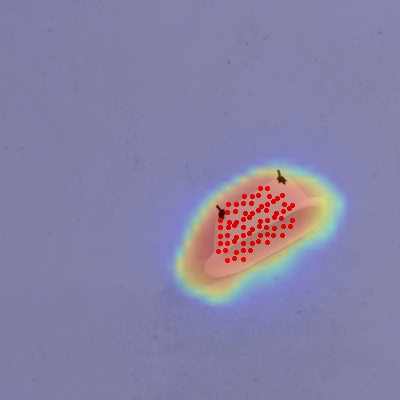}&%
	  \includegraphics[width=0.3\columnwidth]{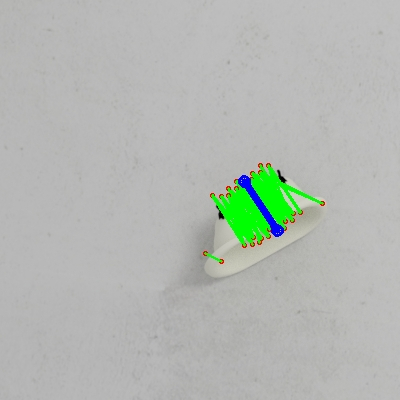}\\
     \includegraphics[width=0.3\columnwidth]{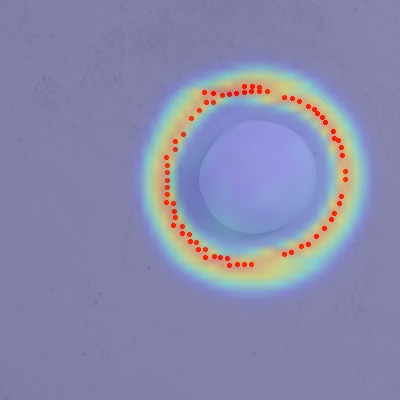}&%
	  \includegraphics[width=0.3\columnwidth]{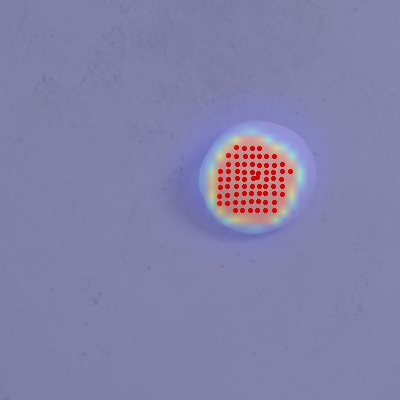}&%
	  \includegraphics[width=0.3\columnwidth]{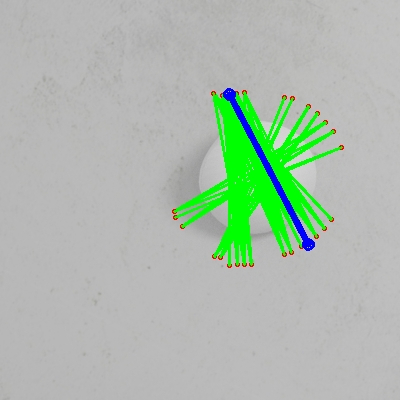}\\
      \includegraphics[width=0.3\columnwidth]{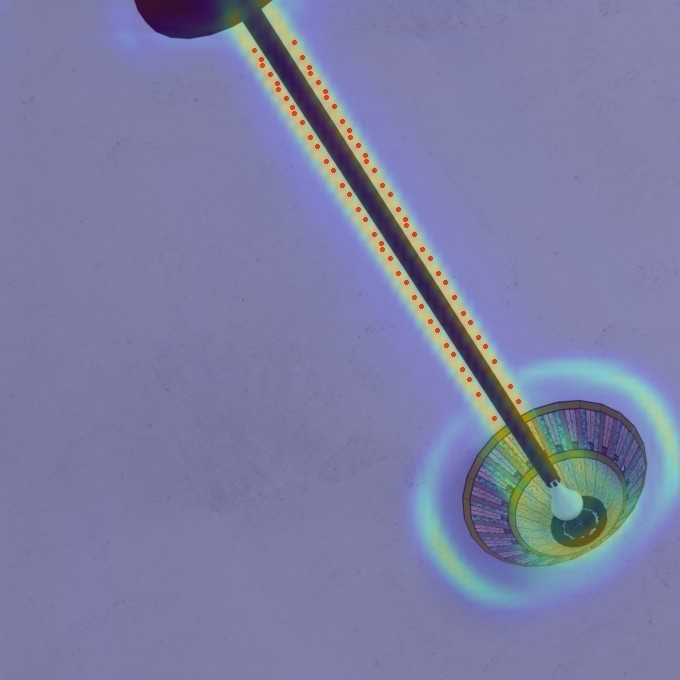}&%
      \includegraphics[width=0.3\columnwidth]{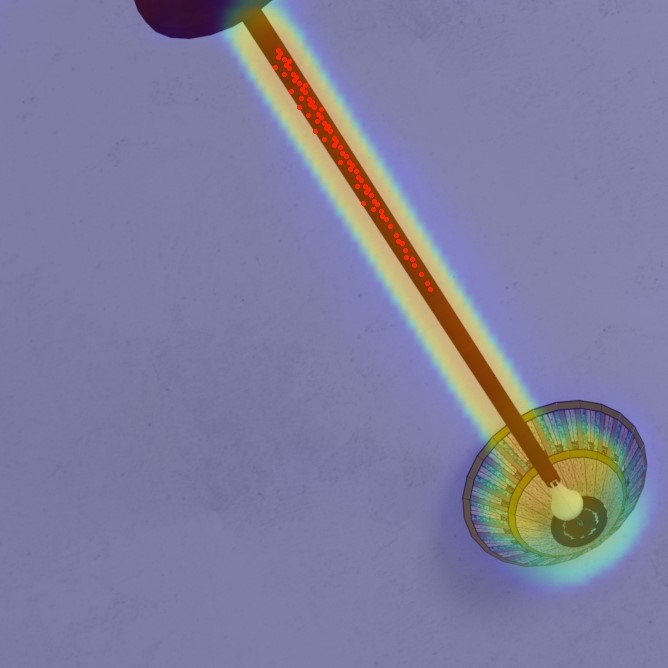}&%
	  \includegraphics[width=0.3\columnwidth]{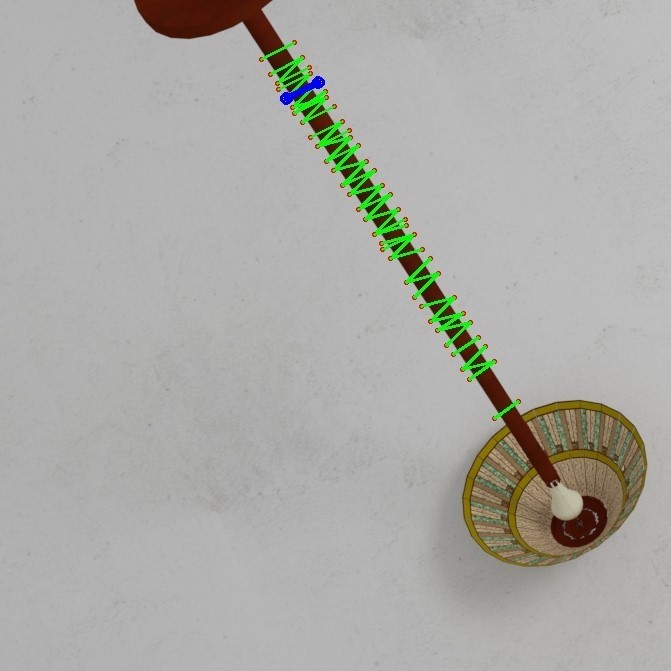}\\
		(a)  & (b)  &(c)  \\
	\end{tabular}
	\caption{Result visualization of DD-Net: (a) the fingertip heatmap and top-k candidates; (b) the center heatmap and top-k candidates; and (c) the fingertip pairs refined by the two matching strategies (blue line indicates final grasp prediction).}
	\label{fig:results}
\end{figure}

\subsection{Further Analysis}

(1) \textbf{Ablation study}: To better understand the major components of DD-Net, we launch an additional experiment to validate the necessity of the orientation matching and center matching strategies in fingertip grouping. The SGT scores are reported as in \autoref{tab:tab12}, and we can see that both the two schemes contribute to performance gain and orientation matching plays a dominant role.

(2) \textbf{Visualization}: Intermediate results are shown in Fig.~\ref{fig:results}. We can see that the locations of fingertip candidates predicted by the network distribute at the surrounding of the objects, which coincide with the expectation of human beings. With the help of the grasp orientation and center, the network selects the best fingertip pair from all the candidates as shown in Fig.~\ref{fig:results} (c).

\vspace{2mm}
\section{Conclusion}
This paper addresses the issue of antipodal grasp detection and formulates it as predicting a pair of fingertips to improve the performance on unseen objects. A novel deep network, DD-Net, is proposed to accurately and robustly localize such points. Furthermore, a specialized loss is designed to measure the grasp quality in training and two matching strategies are introduced to filter fingertip candidates in inference. Both the simulation and robotic experiments are carried out and state-of-the-art results are reached, demonstrating the effectiveness of the proposed approach.

\section*{ACKNOWLEDGMENT}

This work is partly supported by the National Natural Science Foundation of China (No. 62022011), the Research Program of State Key Laboratory of Software Development Environment (SKLSDE-2021ZX-04), and the Fundamental Research Funds for the Central Universities.

\bibliographystyle{IEEEtran}
\bibliography{references}

\end{document}